\documentclass[11pt,a4paper]{article}
\usepackage[hyperref]{acl2020}
\usepackage{times}
\usepackage{latexsym}
\usepackage{graphicx}

\usepackage{microtype}

\aclfinalcopy % Uncomment this line for the final submission
\title{JARVix at SemEval-2022 Task 2: It Takes One to Know One?\\Idiomaticity Detection using Zero and One-Shot Learning}

\author{ Yash Jakhotiya\thanks{\ \ Equal contribution} \thanks{\ \ Corresponding author}\\
  Georgia Institute of Technology \\
  \texttt{yashjakhotiya@gatech.edu}\And
  Vaibhav Kumar\footnotemark[1]\\
  Georgia Institute of Technology \\
  \texttt{vaibhavk@gatech.edu}\AND
  Ashwin Pathak\footnotemark[1]\\
  Georgia Institute of Technology \\
  \texttt{apathak60@gatech.edu}\And
  Raj Shah\footnotemark[1]\\
  Georgia Institute of Technology \\
  \texttt{rajsanjayshah@gatech.edu}\\
  }

\date{}

\begin{document}
\maketitle
\begin{abstract}
Large Language Models have been successful in a wide variety of Natural Language Processing tasks by capturing the compositionality of the text representations. In spite of their great success, these vector representations fail to capture meaning of idiomatic multi-word expressions (MWEs). In this paper, we focus on the detection of idiomatic expressions by using binary classification, based on Subtask A of SemEval-2022 Task 2 \cite{Madabushi-semeval}. Thereafter, we perform the classification in two different settings: zero-shot and one-shot, to determine if a given sentence contains an idiom or not. N shot classification for this task is defined by N number of common idioms between the training and testing sets. In this paper, we train multiple Large Language Models in both the settings and achieve an F1 score (macro) of 0.73 for the zero-shot setting and an F1 score (macro) of 0.85 for the one-shot setting. An implementation of our work can be found at \url{https://github.com/ashwinpathak20/Idiomaticity_Detection_Using_Few_Shot_Learning}.
\end{abstract}
\section{Introduction}
% Talk about task with an example
Transformer-based Large Language Models (LLMs)\cite{large_language_models} like BERT%\cite{BERT}
, DistilBERT%\cite{distilbert}
, RoBERTa and %\cite{roberta}
 their variants show state of art performance on a large number of NLP tasks, yet, they fail to capture multi-word expressions such as idioms. This is because contextualized pre-trained models learn compositional representations of text at sub-word and word level to reduce the vocabulary size. 
 
 Therefore, we evaluate how well do LLMs identify idiomaticty by formulating the problem as a classification task. \\
% The data
\\
In this paper, we propose an approach for Subtask A of SemEval-2022 Task 2 \cite{Madabushi-semeval}. We treat the development data as held-out development data, and report our performance on the test data. To evaluate how well LLMs identify idiomaticity, we use two different settings to determine the generalizability of the LLMs: zero-shot and one-shot setting. The zero-shot setting is defined such that the MWEs in the train set are mutually exclusive of the MWEs found in the test set. For the one-shot setting, there is only one Idiomatic and/or one Literal training example for one MWE in the development set. This is different from traditional definitions of zero-shot and one-shot classification.

The rest of the paper describes the related works in section II and the dataset used in Section III. Section IV gives the methodology used in zero-shot and one-shot learning. Section V describes the performed experiments and Section VI discusses the results. Section VII concludes the paper with a discussion on future research prospects and directions.

\section{Related Work}

Idiomaticity identification for MWEs has been widely studied for single token representation using statistical and semantic methods ~\citep{Lin:99, Baldwin:03}.

Recent works use contextual representations without any token representation for idiomaticity identification for MWEs~\citep{Hashempour:20}. ~\citep{Madabushi:21} introduces new tokens for MWEs into a contextual pre-trained language model. However, they do not explore the relationship of potential MWEs in a sentence.

To this end, we present a contextual and compositional network incorporating latent semantic significance of MWEs in a sentence. Using word embeddings for semantic similarity have been explored before~\citep{Katz:06}. However, the challenge for the semantic usage identifcation of MWEs lies in the ambiguity in meanings of MWEs. Additionally, low frequency occurrences of MWEs inhibit the models to effectively learn the contextual representations as well. 

Siamese Networks have been widely used for similarity detection and difference tracking. We propose to carry forward this idea for identification of idioms in MWEs by comparing the literal usage of MWEs from their idiomatic usage. This enables our approach to learn a contextual and compositional structure within a sentence.

\iffalse 0
MWEs pose a wide range of problems for detection~\citep{Constant:17}. Idiomaticity identification was initially addressed by using statistical properties of text, symbolic methods, and latent semantic analysis.

Further, constituent word embeddings were used for semantic similarity between the distributional vectors associated with an MWE and its parts. These were improved by an explicit disambiguation step~\citep{Kartsaklis:14} and by the joint
learning of compositional and idiomatic embeddings~\citep{Hashimoto:16}. However, these methods have their shortcomings due to low frequency of MWEs and limited non-contextual type level representations of MWEs multiple meanings.

Transformer-based pre-trained models do not benefit MWE identification without a new token representation. We 
\fi

\section{Method}

SemEval 2022 task 2 Subtask A \citep{Madabushi:21, Madabushi-semeval} is a task to evaluate the extent to which models can identify idiomaticity in text through a coarse-grained classification into an ``Idiomatic'' or ``Non-idiomatic'' class. To better evaluate a model’s ability to generalise and learn in a sample efficient fashion, the scores are reported in the zero-shot and one-shot setups.

\paragraph{Data \\}
The dataset used in this report is the one provided by \citep{Madabushi:21}. Each of the train and development splits of this dataset consists of samples containing a target sentence, it's language information, a multiword expression (MWE), two contextual sentences that occur before and after the target sentence, and a label associated with the target. The label represents whether the multiword expression was used in an idiomatic sense or not.

The train split is further divided into zero-shot and one-shot data, containing 4491 and 140 samples each, consisting of 236 and 100 distinct MWEs respectively. Similarly, the development data contains 739 samples made from 50 different MWEs. One-shot MWEs have no overlap with zero-shot ones. However, development data MWEs are a proper subset, as can be expected in a one-shot classification scenario. 

\iffalse
The language distribution of the dataset is summarized in Table \ref{language-distribution}.
\begin{table}
\footnotesize
\centering
\begin{tabular}{cccc}
\hline
\textbf{Language} & \textbf{Split} & \textbf{MWEs} & \textbf{Samples}\\
\hline

English & development & 30 & 466 \\
English & one-shot & 60 & 87 \\
English & zero-shot & 163 & 3327 \\
Portuguese & development & 20 & 273 \\
Portuguese & one-shot & 40 & 53 \\
Portuguese & zero-shot & 73 & 1164 \\
\hline
\end{tabular}
\caption{\label{language-distribution}
Language distribution in the dataset}
\end{table}

Even though all the MWEs present in the development set are present in one-shot data, not all of them have both idiomatic and non-idiomatic samples. 17 out of the 50 development set MWEs have only a non idiomatic sample present in one-shot data, and 12 development set MWEs have only an idiomatic sample present. Additionally, no MWE-Label pair is ever repeated in one-shot data, making this dataset and the task one-shot with respect to distinct MWEs.
\fi

\paragraph{Zero-shot learning \\} For the zero-shot learning task, we use the train data to build a classifier using large language models like BERT-multilingual-uncased, DistilBERT-multilingual-uncased, XLM-RoBERTa-large and XLM-net. This task is ``zero-shot'' in nature as the idioms used in the train set and the development set are distinct. Therefore, we capture the discrepancy in the contextual meaning for idiomaticity, that is, we aim that our classifier distinguishes on the basis of lack of semantic correctness of literal meaning in the presence of an idiom in a sentence.\\
To make sure that idioms are not used explicitly while pre-training in large language models, we run a natural language inference task on BART-Large-MNLI and RoBERTa-Large-MNLI with the hypothesis as ``idiom''. The macro F1 score for both approaches is 0.51 and 0.50 respectively, which proves that there is no semantically learnt concept of ``idiomaticity'' by the model. No training data was used for this step.

We therefore use multilingual LLMs to build classifiers for this setting. We need multilingual classifiers as the data consists of idioms in three languages: English, Portuguese and Galician. We further analyse the majority voting approach on the predictions of trained classifiers (inference based ensembling).

% Furthermore, we try two types of ensembling approaches. The first approach is a majority voting approach on the predictions of trained classifiers (inference based ensembling). The second approach trains multiple models simultaneously to build a better classifier.
\paragraph{One-shot learning \\}

\begin{figure}[htp]
    \centering
    \includegraphics[width=7cm]{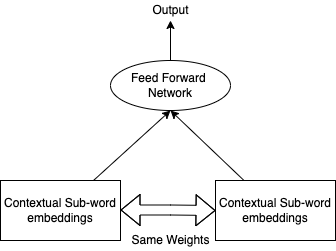}
    \caption{One-shot learning framework}
    \label{fig:galaxy}
\end{figure}

In the one-shot setting, we use the only positive and/or the only negative training example, as available for each MWE in the development set. Note that the actual examples in the training data are different from those in the development set in both settings.

As shown in Fig \ref{fig:galaxy}, our model relies on finding similarity or relation scores between two input sentences. We first train this model on the pretext task of predicting whether two sentences with the same MWE belong to the same class. To achieve this goal, we employ contextual word embeddings to encode two sentences into feature vectors via an embedding function ${f_{\theta}}$. The feature vectors are then combined with an operator $\mathcal{O(.,.)}$ to output $\mathcal{O}(f_{\theta}(x_i),f_{\theta}(x_j))$ on two inputs $x_i$ and $x_j$. This is finally passed to a similarity/relation function $g_{\phi}$ to give score $s_{i,j}$ as,

$$s_{i,j} = g_{\phi}(\mathcal{O}(f_{\theta}(x_i),f_{\theta}(x_j)))$$

We test this framework with two underlying models - a Siamese Neural Network \citep{Koch:15} and a Relation Network \citep{sung2018learning}. With the Siamese Network, the operator $\mathcal{O(.,.)}$ is the element-wise difference between the two input feature vectors. The function $g_{\phi}$ is a fully connected layer followed by sigmoid activation. The loss in this case can be defined as,
${L}(s_{i,j}) = \Sigma_{i,j}1_{y_{i}=y_{j}}\log(s_{i,j})+(1-1_{y_{i}=y_{j}})\log(1-s_{i,j})$, where $y_{i}$ and $y_{j}$ are the labels associated with $x_{i}$ and $x_{j}$. Similarly, for the Relation Network, $\mathcal{O(.,.)}$ becomes the concatenation operator, $g_{\phi}$ becomes three fully connected layers with non linear activations followed by a sigmoid activation function. The loss in this case is the MSE loss, ${L}(s_{i,j}) = \frac{1}{n}\Sigma_{i,j}(s_{i,j} - 1(y_{i}==y_{j}))^2$. In both of the models, $x_i, x_j$ pairs are samples with matching MWEs.

We propose a novel inference methodology for our binary classification problem, where we also consider a dissimilarity score $1 - s_{i,j}$, with $x_i, x_j$ belonging to support and query sets respectively. Support set is defined to be all samples with the same MWE as the query. We find the maximum of similarity and dissimilarity scores for all examples in the support set, and assign the same label or the opposite depending on whether the maximum was the similarity or the dissimilarity score. This helps us with \citep{Madabushi:21} dataset where one-shot training data doesn't have samples for both the classes (idiomatic and non-idiomatic) for all MWEs.     
\section{Experiments}

\paragraph{Zero-shot learning \\}
We run our experiments on pre-trained models for zero-shot classification. We use Multilingual BERT, Multilingual DistilBERT, BERT-Portuguese, XL-Net and XLM-RoBERTa for exhaustive comparison and evaluation. We ensemble XL-NET, XLM-RoBERTa, and Multilingual DistilBERT in a majority vote based setting. As per the SemEval task, our baseline is Multilingual BERT for classification.

\paragraph{One-shot learning \\}
For the contextual embeddings, we run our experiments on pre-trained compositional multi-lingual base models BERT, DistilBERT and XLM-RoBERTa for exhaustive comparison and evaluation. We run Siamese networks with cross entropy loss and Relation Networks with an MSE loss. 

Our hyperparameter search pointed towards a dropout rate of 0.5, a learning rate of 2e-5 and we found AdamW to be the best performing optimizer. 

\section{Results}

\begin{table}[htb]
\centering
\begin{tabular}{ccc}
\hline
\textbf{LN} & \textbf{Model} & \textbf{Dev F1}\\
\hline
EN & BERT & 0.65 \\
EN & DistilBERT & 0.70 \\
EN & XLM-RoBERTa & \textbf{0.73} \\
EN & XL-NET & \textbf{0.73} \\
EN & Ensemble & 0.71 \\
\hline
PT & BERT & \textbf{0.64} \\
PT & DistilBERT & 0.58 \\
PT & XLM-RoBERTa & 0.63 \\
PT & XL-NET & 0.62 \\
PT & Ensemble & 0.53 \\
\hline
EN-PT & BERT & 0.67 \\
EN-PT & DistilBERT & 0.70 \\
EN-PT & XLM-RoBERTa & 0.71 \\
EN-PT & XL-NET & \textbf{0.73} \\
EN-PT & Ensemble & 0.68 \\
\hline
\end{tabular}
\caption{\label{citation-guide}
Zero-shot evaluation results}
\label{tab:results}
\end{table}

\begin{table}[htb]
\footnotesize
\centering
\begin{tabular}{cccc}
\hline
\textbf{LN} & \textbf{Emb Model} & \textbf{Siamese F1} & \textbf{Relation F1}\\
\hline
EN &  BERT & 0.79 & \textbf{0.85}\\
EN &  DistilBERT & 0.79 & 0.83\\
EN &  XLM-RoBERTa & 0.83 & \textbf{0.85}\\
\hline
PT & BERT & 0.81 & 0.84\\
PT & DistilBERT & 0.80 & \textbf{0.85}\\
PT & XLM-RoBERTa & 0.85 & \textbf{0.85}\\
\hline
EN-PT & BERT & 0.80 & \textbf{0.85}\\
EN-PT & DistilBERT & 0.79 & 0.84\\
EN-PT & XLM-RoBERTa & 0.84 & \textbf{0.85}\\
\hline
\end{tabular}
\caption{\label{citation-guide}
One-shot evaluation results}
\label{tab:results-1-shot}
\end{table}

\begin{table}[htb]
\centering
\begin{tabular}{ccc}
\hline
\textbf{Setting} & \textbf{Language} & \textbf{Test F1}\\
\hline
Zero-shot & EN & 0.7869 \\
Zero-shot & PT & 0.7201 \\
Zero-shot & GL & 0.5588 \\
Zero-shot & EN,PT,GL & 0.7235 \\
\hline
One-shot & EN & 0.8410 \\
One-shot & PT & 0.8162 \\
One-shot & GL & 0.7918 \\
One-shot & EN,PT,GL & 0.8243 \\
\hline
\end{tabular}
\caption{\label{citation-guide}
Test evaluation results}
\label{tab:results-test}
\end{table}

\paragraph{Zero-shot learning \\}
Table \ref{tab:results} shows F1-scores for different configurations, both ensemble and individual language models, with the baseline model being Multilingual BERT. We observe that the ensemble model performs better than the baseline in case of EN (0.71 F1 score) and EN-PT (0.68 F1 score) as compared to PT (0.53 F1 score) data. We further observe that XL-NET outperforms other models in case of English and Portuguese inputs. Our best performing zero-shot setting results in a 0.72 F1 score on the test split of the dataset,, which is a significant boost from the 0.65 F1 score in the baseline provided by \cite{Madabushi:21}.

\paragraph{One-shot learning \\}

Table \ref{tab:results-1-shot} reports F-1 scores for one-shot learning. We found the best results of our Siamese and Relation network with XLM-RoBERTa (0.85 F1-score). We also observed a better score for Portuguese dataset as compared to English dataset on all of our models. Our best performing relation networks get 0.82 F1 score on the test split, which is competitive with \cite{Madabushi:21}.

Table \ref{tab:results-test} breaks down our test set evaluation results by language. GL in the table stands for Galician, which had data only in the test split.

\section{Analysis and Conclusion}
In this paper we analyzed the effectiveness of large Language Models towards identifying idiomaticity in a given phrase using zero-shot and one-shot classification tasks. 
% Our experiments highlight the efficacy of introducing a new token representation for MWE as input to the pre-trained language model \cite{Madabushi:21}. 

In zero-shot classification, we use inference-level ensembling of different language models and observe that it outperforms BERT baseline in cases where the input language consists of English. This highlights a high degree of disagreement amongst the language models w.r.t Portuguese input, highlighting their brittleness. 

For one-shot classification, through Siamese and Relation Networks, we are able to represent the latent semantic relationship among MWEs leading to a much better F1 score than zero-shot classification and competitive with prior work. We believe that the improvement in performance of the relation network comes due to the learn-able nature of the distance function used between query and support data sample, as well as our novel inference methodology which also takes into account the dissimilarity score. Future work for one-shot classification could aim at breaking the barrier of 0.85 F1 score we seem to have hit on the dev set with all embedding base models.

\bibliography{anthology,acl2020}

\begin{thebibliography}{9}
\expandafter\ifx\csname natexlab\endcsname\relax\def\natexlab#1{#1}\fi

\bibitem[{Baldwin and Villavicencio(2002)}]{Baldwin:03}
Timothy Baldwin and Aline Villavicencio. 2002.
\newblock \href {https://aclanthology.org/W02-2001} {Extracting the
  unextractable: A case study on verb-particles}.
\newblock In \emph{{COLING}-02: The 6th Conference on Natural Language Learning
  2002 ({C}o{NLL}-2002)}.

\bibitem[{Hashempour and Villavicencio(2020)}]{Hashempour:20}
Reyhaneh Hashempour and Aline Villavicencio. 2020.
\newblock Leveraging contextual embeddings and idiom principle for detecting
  idiomaticity in potentially idiomatic expressions.
\newblock \emph{In Proceedings of the Workshop on the Cognitive Aspects of the
  Lexicon}, pages 72--80.

\bibitem[{Kant et~al.(2018)Kant, Puri, Yakovenko, and
  Catanzaro}]{large_language_models}
Neel Kant, Raul Puri, Nikolai Yakovenko, and Bryan Catanzaro. 2018.
\newblock \href {http://arxiv.org/abs/1812.01207} {Practical text
  classification with large pre-trained language models}.
\newblock \emph{CoRR}, abs/1812.01207.

\bibitem[{Katz and Giesbrecht(2006)}]{Katz:06}
Graham Katz and Eugenie Giesbrecht. 2006.
\newblock Automatic identification of non-compositional multiword expressions
  using latent semantic analysis.
\newblock \emph{In Proceedings of the Workshop on Multiword Expressions}, pages
  12--19.

\bibitem[{Koch et~al.(2015)Koch, Zemel, and Salakhutdinov}]{Koch:15}
Gregory Koch, Richard Zemel, and Ruslan Salakhutdinov. 2015.
\newblock Siamese neural networks for one-shot image recognition.
\newblock \emph{In Proceedings of the 32nd International Conference on Machine
  Learning}.

\bibitem[{Lin(1999)}]{Lin:99}
Dekang Lin. 1999.
\newblock Automatic identification of noncompositional phrases.
\newblock \emph{In Proceedings of the 37th Annual Meeting of the Association
  for Computational Linguistics}, pages 317–--324.

\bibitem[{Sung et~al.(2018)Sung, Yang, Zhang, Xiang, Torr, and
  Hospedales}]{sung2018learning}
Flood Sung, Yongxin Yang, Li~Zhang, Tao Xiang, Philip H.~S. Torr, and
  Timothy~M. Hospedales. 2018.
\newblock \href {http://arxiv.org/abs/1711.06025} {Learning to compare:
  Relation network for few-shot learning}.

\bibitem[{Tayyar~Madabushi et~al.(2022)Tayyar~Madabushi, Gow-Smith, Garcia,
  Scarton, Idiart, and Villavicencio}]{Madabushi-semeval}
Harish Tayyar~Madabushi, Edward Gow-Smith, Marcos Garcia, Carolina Scarton,
  Marco Idiart, and Aline Villavicencio. 2022.
\newblock {SemEval-2022 Task 2}: {Multilingual Idiomaticity Detection and
  Sentence Embedding}.
\newblock In \emph{Proceedings of the 16th International Workshop on Semantic
  Evaluation (SemEval-2022)}. Association for Computational Linguistics.

\bibitem[{Tayyar~Madabushi et~al.(2021)Tayyar~Madabushi, Gow-Smith, Scarton,
  and Villavicencio}]{Madabushi:21}
Harish Tayyar~Madabushi, Edward Gow-Smith, Carolina Scarton, and Aline
  Villavicencio. 2021.
\newblock \href {https://doi.org/10.18653/v1/2021.findings-emnlp.294}
  {{AS}titch{I}n{L}anguage{M}odels: Dataset and methods for the exploration of
  idiomaticity in pre-trained language models}.
\newblock In \emph{Findings of the Association for Computational Linguistics:
  EMNLP 2021}, pages 3464--3477, Punta Cana, Dominican Republic. Association
  for Computational Linguistics.

\end{thebibliography}
\bibliographystyle{acl_natbib}

\end{document}